\documentclass[letterpaper]{article} % DO NOT CHANGE THIS
\usepackage{aaai25}  % DO NOT CHANGE THIS
\usepackage{times}  % DO NOT CHANGE THIS
\usepackage{helvet}  % DO NOT CHANGE THIS
\usepackage{courier}  % DO NOT CHANGE THIS
\usepackage[hyphens]{url}  % DO NOT CHANGE THIS
\usepackage{graphicx} % DO NOT CHANGE THIS
\usepackage{natbib}  % DO NOT CHANGE THIS AND DO NOT ADD ANY OPTIONS TO IT
\usepackage{caption} % DO NOT CHANGE THIS AND DO NOT ADD ANY OPTIONS TO IT
\usepackage{bm}
\usepackage{algorithm}
\usepackage{algorithmic}
\usepackage{multirow}
\usepackage{expl3}
\usepackage{amsmath}
\usepackage{amssymb}
\usepackage[usenames,dvipsnames]{color}

\newcommand{\eg}{\textit{e.g.}}

\usepackage{newfloat}
\usepackage{listings}
\DeclareCaptionStyle{ruled}{labelfont=normalfont,labelsep=colon,strut=off} % DO NOT CHANGE THIS
\floatstyle{ruled}
\newfloat{listing}{tb}{lst}{}
\floatname{listing}{Listing}

\title{ERL-MPP: Evolutionary Reinforcement Learning with Multi-head Puzzle Perception  for Solving Large-scale Jigsaw Puzzles of Eroded Gaps}
\author{
    Xingke Song\textsuperscript{\rm 1},
    Xiaoying Yang\textsuperscript{\rm 1},
    Chenglin Yao\textsuperscript{\rm 1},
    Jianfeng Ren\textsuperscript{\rm 1}\textsuperscript{\rm 2}\thanks{Corresponding Authors.},
    Ruibin Bai\textsuperscript{\rm 1}\textsuperscript{\rm 2},
    Xin Chen\textsuperscript{\rm 3},
    Xudong Jiang\textsuperscript{\rm 4},
}
\affiliations{
    \textsuperscript{\rm 1}School of Computer Science, University of Nottingham Ningbo China, China\\
    \textsuperscript{\rm 2}Nottingham Ningbo China Beacons of Excellence Research and Innovation Institute, \\University of Nottingham Ningbo China, China\\
    \textsuperscript{\rm 3}Intelligent Modelling \& Analysis Group, School of Computer Science, University of Nottingham, Nottingham, UK\\
    \textsuperscript{\rm 4}School of Electrical \& Electronic Engineering, Nanyang Technological University, Singapore\\
    \{Xingke.Song, Xiaoying.Yang, Chenglin.Yao, Jianfeng.Ren, Ruibin.Bai\}@nottingham.edu.cn, Xin.Chen@nottingham.ac.uk, EXDJiang@ntu.edu.sg
}

\usepackage{bibentry}
\begin{document}

\maketitle

\begin{abstract} 
Solving jigsaw puzzles has been extensively studied. While most existing models focus on solving either small-scale puzzles or puzzles with no gap between fragments, solving large-scale puzzles with gaps presents distinctive challenges in both image understanding and combinatorial optimization. To tackle these challenges, we propose a framework of Evolutionary Reinforcement Learning with Multi-head Puzzle Perception (ERL-MPP) to derive a better set of swapping actions for solving the puzzles. Specifically, to tackle the challenges of perceiving the puzzle with gaps, a Multi-head Puzzle Perception Network (MPPN) with a shared encoder is designed, where multiple puzzlet heads comprehensively perceive the local assembly status, and a discriminator head provides a global assessment of the puzzle. To explore the large swapping action space efficiently, an Evolutionary Reinforcement Learning (EvoRL) agent is designed, where an actor recommends a set of suitable swapping actions from a large action space based on the perceived puzzle status, a critic updates the actor using the estimated rewards and the puzzle status, and an evaluator coupled with evolutionary strategies evolves the actions aligning with the historical assembly experience. The proposed ERL-MPP is comprehensively evaluated on the JPLEG-5 dataset with large gaps and the MIT dataset with large-scale puzzles. It significantly outperforms all state-of-the-art models on both datasets.  
\end{abstract}

\section{Introduction}
\label{sec: intro}
Jigsaw puzzles are popular entertainment and intellectual challenges. Solving jigsaw puzzles has been related to a wide range of applications, \emph{e.g.}, geometric analysis~\cite{Harel2024Pictorial}, boundary information understanding~\cite{gallagher2012jigsaw}, and image semantic understanding~\cite{song2023sd2rl}. In particular, automatic puzzle solving has been extensively studied in archaeology~\cite{Rasheed2015ASO} and cultural heritage restoration~\cite{Derech2021solving}, which greatly releases the burden of archaeologists. The developed techniques have been applied beyond puzzle reassembly, \emph{e.g.}, self-supervised learning~\cite{doersch2015unsupervised,noroozi2016unsupervised,ren2023masked}, video spatial understanding~\cite{Yang2022p2pnet}, fine-grained visual classification~\cite{du2020fine}, and image super-resolution~\cite{ma2021sr}.
\begin{figure}[!t]
    \centering
    \includegraphics[width=1.\columnwidth]{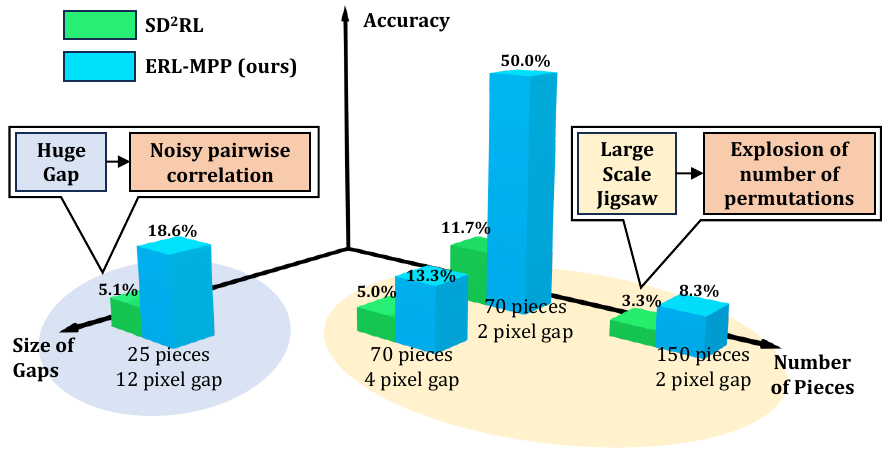}
    \caption{
    Huge gaps and a large amount of fragments impose great challenges on puzzle solvers. The proposed ERL-MPP significantly outperforms the previous best performing SD$^2$RL~\cite{song2023sd2rl} on four different types of puzzles, \eg, puzzles with as large as 12-pixel gaps, or puzzles with as many as 150 pieces of fragments.}
    \label{fig:intro}
\end{figure}

Most methods focus on solving puzzles with no gaps or very small gaps between fragments~\cite{son2019solving,bridger2020solving,Li2022JigsawGAN}, and utilize the boundary information to infer the pairwise relations between fragments. However, in real-world applications, there are often gaps on artifacts due to damage or erosion~\cite{paumard2020deepzzle,song2023sd2rl}, making the boundary similarity unavailable. This paper tackles the problem of solving Large-scale Jigsaw Puzzles of Eroded Gaps (L-JPEG). Challenges are two-fold.    
1) Visual perception of fragments under gaps.  Large gaps make the methods based on the boundary information ineffective, resulting in noisy estimation of relations between adjacent fragments.  
2) A huge search space for optimization strategy. The number of permutations increases exponentially with the number of puzzle pieces, coupled with the uncertainty in pairwise relations, making the puzzle reassembly computationally demanding.

In this paper, we propose a framework of Evolutionary Reinforcement Learning with Multi-head Puzzle Perception (ERL-MPP) to derive a better set of swapping actions to assemble the puzzle, leading to the significant performance gain as shown in Fig.~\ref{fig:intro}. Firstly, to perceive visual clues from puzzles effectively, a Multi-head Puzzle Perception Network (MPPN) is designed as the perception module, where three puzzlet perception heads excavate the local features at multiple scales by comparing fragments in local neighborhoods, and a discriminator head globally evaluates whether the puzzle has been correctly reassembled. The feature encoder, decoder, and discriminator form a Generative Adversarial Network (GAN), where the discriminator evaluates the quality of decoded image as a reassembled puzzle and helps refine the encoder to capture the critical features in assessing the puzzle status. The puzzlet perception heads and the discriminator head share the same feature encoder to not only reduce the network complexity but also produce robust features.

Secondly, to tackle the challenges of assembling a large number of puzzle pieces, an Evolutionary Reinforcement Learning (EvoRL) agent is designed to select a set of fragment-swapping actions based on both the current visual perception and the past reassembly experience stored in the agent.  
The proposed EvoRL consists of an actor, a critic, and an evaluator. While the actor predicts the probability of feasible actions from a given action space, the critic evaluates the current puzzle status and updates the actor based on the visual perception from the MPPN and the reward collected from the environment. The evaluator assesses the quality of actions evolved using techniques such as crossover with the action population from past reassembly experience and natural mutations, which helps discover high-performance actions, facilitating the agent to fast converge to an optimal solution. Besides the actions of swapping two fragments~\cite{song2023sd2rl}, the agent explores the actions of swapping three fragments and swapping two $2\times2$ puzzlets. During training, the agent initially exploits larger size swapping actions to explore the action space more efficiently with the help of the evolutionary strategies to avoid a local optimum, and gradually utilizes smaller size swapping actions to focus on assembling the puzzle perfectly. A set of rewards are designed to provide clear training guidance for the agent, promoting the correct fragment placement and perfect puzzle reassembly. By utilizing the multiple perceptions of the MPPN and the effective exploration of the EvoRL, the proposed model better explores and exploits the large action space to derive a better set of swapping actions to perfectly reassemble the puzzle. 

Our contributions can be summarized as follows.  
1)~The proposed MPPN perceives both locally adjacent fragments in small neighborhood through the puzzlet perception heads and the global puzzle status through the discriminator head.
2)~The proposed EvoRL parameterizes the action probabilities from the action space and utilizes the evolutionary strategies to evolve series of best swapping trajectories, alleviating the problem of enumerating and evaluating all possible swapping actions in the value-based RL agents~\cite{song2023sd2rl}, which effectively explores the large action space for large-scale puzzles.
3)~The proposed ERL-MPP consistently and significantly outperforms the state-of-the-art models on the JPLEG-5 dataset~\cite{song2023sd2rl} with large gaps and the MIT dataset~\cite{bridger2020solving} with large-scale puzzles. 

\begin{figure*}[!t]
    \centering
    \centerline{\includegraphics[width=1\linewidth]{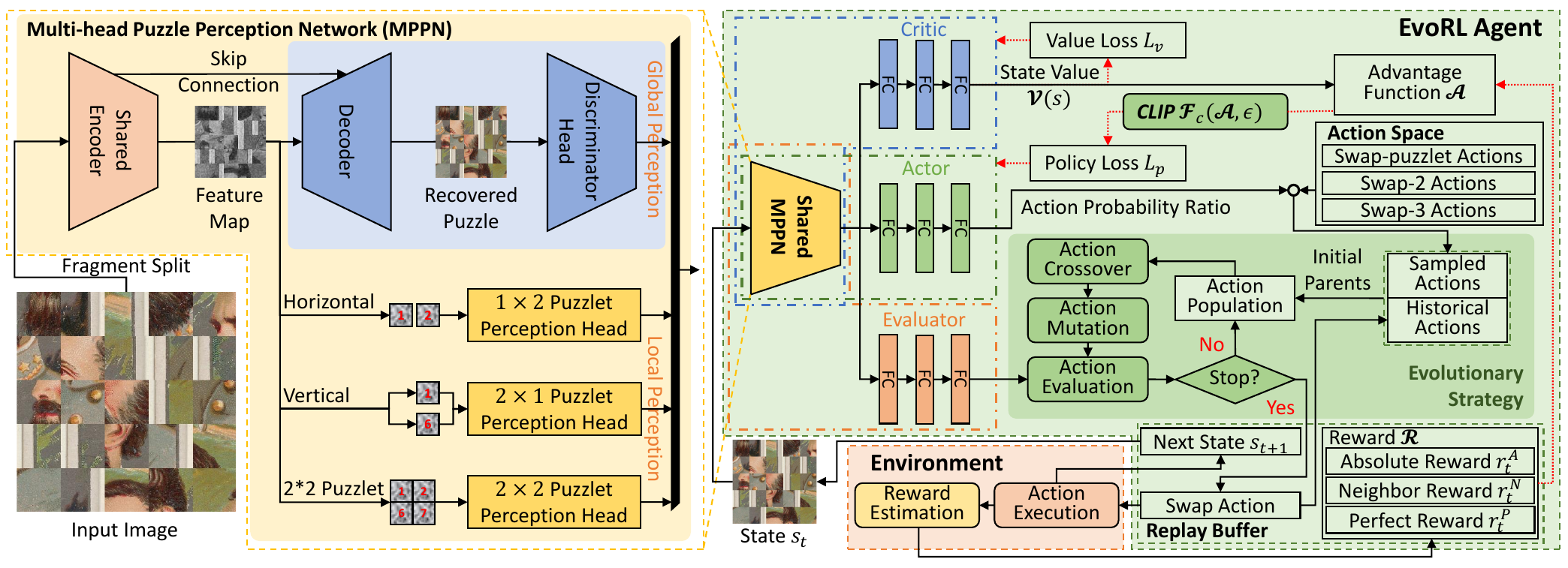}}
    \caption{Block diagram of the proposed ERL-MPP for solving L-JPEG problems. The shared MPPN globally perceives the puzzle through a discriminator head and locally perceives it through three puzzlet perception heads. The EvoRL agent determines an optimal sequence of swapping actions till perfectly reassembling the puzzle, where an actor recommends swapping actions from a large action space of \texttt{Swap-2}, \texttt{Swap-3}, and \texttt{Swap-Puzzlet} actions, a critic estimates the state value and updates the actor based on the visual perception from the MPPN and the estimated reward, and an evaluator assesses and selects the most suitable action after evolutionary operations such as crossover and mutation. A set of rewards considering fragment placement, pairwise adjacency and perfect reassembly are designed to guide the training of the agent.} 
    \label{fig:PPO}
\end{figure*}

\section{Related Work}
\label{sec: related-work}
Puzzle solving often involves visual perception of puzzles and puzzle reassembly based on the perceived information.

\noindent\textbf{Visual Perception of Puzzles.}
Traditional methods mainly handle puzzles with no gap or very small gaps between fragments, and utilize the boundary information to reliably estimate the pairwise relations between fragments, \eg, Handcrafted features such as boundary shapes~\cite{zhang20153d}, color contrast~\cite{son2019solving,Yan2021Solving}, and boundary similarities~\cite{sholomon2013genetic,paikin2015solving} are often utilized to describe adjacent fragment pairs. Deep neural networks have also been developed to extract boundary features~\cite{Khoroshiltseva2022jigan,Li2022JigsawGAN}.

Recent studies focus on solving puzzles with gaps, where the boundary information is unavailable due to relatively large gaps, \emph{e.g.}, 2-4 pixels in~\cite{bridger2020solving}, 12 pixels in~\cite{song2023solving,song2023sd2rl} and 48 pixels in~\cite{paumard2020deepzzle,song2023sd2rl,song2023solving}. In this case, image semantics are often utilized to exploit pairwise relations~\cite{paumard2020deepzzle,song2023sd2rl,song2023solving}. 
In~\cite{bridger2020solving,Khoroshiltseva2022jigan}, the correlation between adjacent pieces is estimated by inpainting the small gaps using GAN. But it is difficult to inpaint the large gaps of 12 pixels or 48 pixels. The image semantics are extracted in this case using deep neural networks to estimate the adjacency relations~\cite{paumard2018image,paumard2020deepzzle,song2023sd2rl,song2023solving}. To predict the correlation between the central fragment and neighboring fragments, \citet{paumard2020deepzzle} developed a Siamese Neural Network. In SD$^2$RL, \citet{song2023sd2rl} designed a set of Siamese Discriminant Networks to infer the pairwise relations. Besides pairwise relations, puzzlets formed by more adjacent pieces are evaluated by Puzzlet Discriminant Network~\cite{song2023solving}. These existing methods often focus on local pairwise features, while global puzzle perception could also help. This paper presents a multi-head puzzle perception network to perceive both local and global puzzle placement status. 

\noindent\textbf{Puzzle Reassembly.} 
Puzzle reassembly strategies can be broadly divided into two groups. 
1) Methods for solving puzzles with no gaps or small gaps~\cite{son2019solving,bridger2020solving}, where adjacency between fragments can be reliably estimated, and maximizing adjacency generally leads to correct placements~\cite{cho2010probabilistic,gallagher2012jigsaw}. These methods focus on accelerating the solution of large-scale puzzles, \emph{e.g.}, greedy tree-based method~\cite{bridger2020solving}, genetic algorithm~\cite{sholomon2013genetic}, and loop constraints~\cite{son2014solving,son2019solving}. 
2) Methods for solving puzzles with large gaps~\cite{paumard2020deepzzle,song2023sd2rl,song2023solving}, where the adjacency between fragments can't be reliably estimated due to the gaps. \citet{paumard2020deepzzle} formulated puzzle reassembly as a shortest-path optimization problem, and solved it by Dijkstra's algorithm with branch-cut. PDN-GA~\cite{song2023solving} utilizes a genetic algorithm to derive the reassembly sequence. \citet{song2023sd2rl} developed a Deep Q-network for puzzle solving, but it is difficult to evaluate all the feasible actions, especially when the puzzle has many fragments. 

\section{Proposed ERL-MPP}
\label{sec: methodology}
% \subsection{Motivations of Proposed Method}
\subsection{Overview of Proposed ERL-MPP}
\label{sec: overallFramework}
The proposed ERL-MPP tackles the L-JPEG problems, where both the gaps and the large number of puzzle pieces pose great challenges. The block diagram is shown in Fig.~\ref{fig:PPO}. The proposed shared MPPN consists of a GAN-like network head to perceive the global puzzle status and three puzzlet perception heads to perceive local puzzle status. The discriminator head and the puzzlet perception heads share the same encoder to generate a comprehensive representation through joint feature learning. The encoder-decoder architecture together with the discriminator is designed to derive robust features focusing on the image semantics. 

The proposed EvoRL agent includes three major components. 
1) An actor to estimate the probability ratio of swapping actions in a given action space, and recommend candidate actions for the evolutionary algorithm to form high quality swap-based search trajectories with the assistance of historical experience. 
2) A critic to estimate the state value based on the visual perception from the MPPN and the reward evaluated from the puzzle placement, 
and update the critic through the critic loss and the actor through the policy loss. 
3) An evaluator to assess the swapping actions in a rollout fashion after crossover and mutation, and to select the most suitable one. It shares the same architecture as the critic and is soft-updated by the critic~\cite{wei2021deep}, learning from the critic while maintaining independence to prevent potential overfitting. The evolution helps explore a wider range of combinations of swapping actions, potentially discovering high-performing actions that the actor might not find on its own.  
Compared with value-based reinforcement learning~\cite{song2023sd2rl}, the EvoRL better balances the exploration and exploitation in a large action space.  

\subsection{Multi-head Puzzle Perception Network}
\begin{figure}[!t]
\centering
\includegraphics[width=1\linewidth]{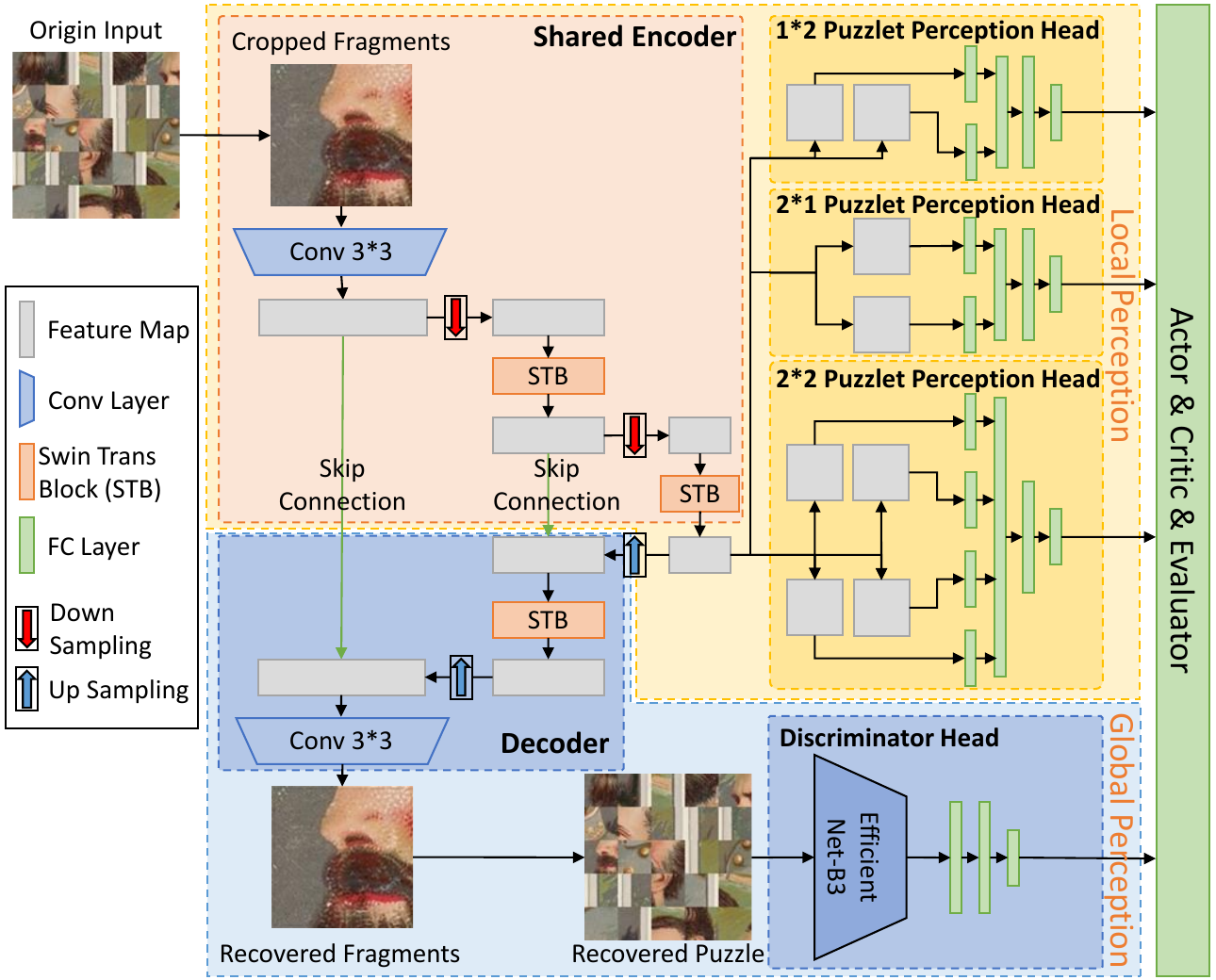}
\caption{The proposed MPPN with a shared encoder, where a discriminator head perceives global puzzle semantics, and puzzlet perception heads perceive local adjacency relations.}
\label{fig:GAN}
\end{figure}

As shown in Fig.~\ref{fig:GAN}, the proposed MPPN consists of four main building blocks: A shared encoder to extract common features, a decoder to generate a fragment based on the features, a discriminator head to globally perceive the puzzle and three puzzlet perception heads to perceive the puzzlets. 

\noindent \textbf{Encoder and Decoder.} 
The encoder and the decoder form a U-shape network architecture~\cite{liu2021swin}, where the convolutional layers are placed at the beginning of the encoder and the end of the decoder to focus on low-level image details, and the Swin Transformer blocks replace the convolution layers in the middle of traditional U-net for feature extraction and pattern recovery at various scales, modeling high-level complex patterns with low computational complexity. Following~\cite{cao2022swin}, each Swin Transformer block contains four Swin Transformer Layers (STLs), and each STL consists of one window multi-head self-attention, and one shifted-window multi-head self-attention, allowing the STLs to capture low-level image fine details and high-level image semantics simultaneously~\cite{cao2022swin}. 

\noindent \textbf{Discriminator.} 
The encoder-decoder architecture faces the difficulty of capturing intricate image characteristic and maintaining visual coherence, while a discriminator could help by contrasting the generated fragments with the real ones~\cite{goodfellow2014gan}. Different from the discriminators applied on puzzle fragments in other models~\cite{bridger2020solving,Khoroshiltseva2022jigan}, our discriminator is applied to the recovered puzzle to assess its global semantic features. It is adapted from the EfficientNet-B3~\cite{tan2019efficientnet} in view of its strong discriminant power, containing 26 MBConv (mobile inverted bottleneck convolution) blocks. Each block consists of a set of convolutional layers and squeeze-and-excitation blocks. The extracted features are fed into three fully connected (FC) layers to aggregate the global image semantics. 

\noindent\textbf{Puzzlet Perception Heads.} 
Three puzzlet perception heads are designed to perceive the horizontal and vertical relations between adjacent fragments and the $2\times 2$ puzzlets.
As shown in Fig.~\ref{fig:GAN}, the heads for horizontal/vertical pairwise relations take the features of two fragments as the input and evaluate their likelihood of being neighbors by aggregating the adjacency information using three fully connected layers. The puzzlet perception head for $2\times 2$ puzzlets takes the features of four fragments as the input to evaluate the likelihood of the puzzlet being correctly assembled or not. 
All the horizontal neighbors, vertical neighbors, and $2\times 2$ puzzlets are evaluated to derive the local perception of puzzles. They are then aggregated together with the global perception from the discriminator to provide visual clues of the current puzzle. 

\noindent\textbf{Aggregation of Visual Clues.} 
Given a puzzle of $M \times N$ fragments, there are $M\times(N-1)$ pieces of $1\times 2$ puzzlets, $(M-1)\times N$ pieces of  $2\times 1$ puzzlets, and $(M-1)\times(N-1)$ pieces of $2\times 2$ puzzlets. To aggregate all the visual clues, we maximize the evidence $E$, 
\begin{align}
    E = \lambda_{g} E_{g} + E_{p},
    \label{eqn:evidence}
\end{align}
where $E_{g}$ is global perception from the discriminator, $E_{p}$ is the perception from the three puzzlet perception heads, and $\lambda_{g}$ is a weighting factor. 
$E_{p}$ is in turn estimated as, 
\begin{align}
E_{p} &= E_{1\times 2} + E_{2\times 1} + E_{2\times 2}, \\
E_{i\times j} &= \sum_{x=1}^{M-i+1} \sum_{y=1}^{N-j+1} \lambda_{i\times j}(x,y) E_{i \times j}(x,y),
\label{eqn:sum_local}
\end{align}
where $i, j \in \{1,2\}$ are the size of puzzlets, $(x,y)$ denotes the location of puzzlets, $\lambda_{i\times j}(x,y)$ is a weighting factor, and $E_{i\times j}$ is the evidence of puzzlets of size $i\times j$.

\subsection{Formulation of Markov Decision Process}
Puzzle reassembly can be formulated as a combinatorial optimization problem~\cite{song2023sd2rl}. Specifically, let $\Pi$ denote the set of permutations of fragment indices $\{1, 2, 3, \dots, MN\}$  for the puzzle of $M \times N$ pieces. 
Given an initial permutation $\pi_0 \in \Pi$, the target is to find a mapping function $\mathcal{M}$ so that $\pi_G =\mathcal{M}(\pi_0)$, where $\pi_G$ is the correct puzzle placement. 
It is difficult to derive such a mapping $\mathcal{M}$ directly. The target is modified to derive a sequence of mappings $\{\mathcal{M}_1, \mathcal{M}_2, \dots \}$, \eg, a sequence of swapping actions $\pi_{t+1} = \mathcal{M}_t(\pi_{t})$ 
that maximize the evidence $E$ defined in Eq.~\eqref{eqn:evidence}, so that the final permutation reaches $\pi_G$. Puzzle reassembly is then formulated as a Markov Decision Process (MDP), represented by the tuple $(\mathbb{S}, \mathbb{A}, \mathcal{R}, P_s)$.

\noindent \textbf{State Space} $\mathbb{S}$, where $s_t \in \mathbb{S}$ is the current state of the puzzle with permutation $\pi_t$.
    
\noindent \textbf{Action Space} $\mathbb{A}$, where $a_t \in \mathbb{A}$ is the action to be applied on the puzzle. Besides common \texttt{Swap-2} actions~\cite{song2023sd2rl}, two more exploratory actions are designed to expand the action space: \texttt{Swap-3} actions that swap three fragments simultaneously, and \texttt{Swap-puzzlet} actions that swap two $2\times 2$ puzzlets. The two new types of swapping actions allow the agent to manipulate more puzzle pieces simultaneously, potentially enabling the agent to escape from local optima and converge fast, while \texttt{Swap-2} actions allow fine adjustments and refinements of puzzles, crucial in the final stages to perfectly reassemble the puzzles. 

\noindent \textbf{Reward} $\mathcal{R}(s_t,a_t)$ for action $a_t$ in state $s_t$ is defined as, 
\begin{align}
\mathcal{R}(s_t,a_t) = \alpha r_t^A + (1-\alpha)r_t^N + r_t^P + b,
\label{eqn:reward}
\end{align} 
where $r_t^A$ rewards the number of fragments being placed at their \textbf{Absolute} locations, $r_t^N$ rewards the number of fragment pairs with correct \textbf{Neighboring} positions, $r_t^P$ is a large reward for the \textbf{Perfect} placement of all fragments and zero otherwise, and $b$ is a small penalty for not correctly reassembling the puzzle at this step. The large reward $r_t^P$ encourages the agent to stop immediately when the puzzle is correctly reassembled, and the small penalty $b$ encourages the agent to converge fast to avoid the lucky success for a long trial. $\alpha \in [0,1]$ balances the importance of $r_t^A$ and $r_t^N$.

\noindent \textbf{Transition Probability} $P_s(s_{t+1}|s_t)$ is defined as the probability that the agent moves from state $s_t$ to state $s_{t+1}$.

\subsection{EvoRL for Puzzle Reassembly}
Existing methods face challenges of handling exponentially increasing search space for L-JPEG problems~\cite{song2023sd2rl,bridger2020solving}. In view of its balanced performance between sample efficiency and solution quality~\cite{martinez2021adaptive}, an Evolutionary Reinforcement Learning agent is developed for solving L-JPEG problems. 
As shown in Fig.~\ref{fig:PPO}, the agent consists of an actor to recommend the swapping actions from a large action space, a critic to estimate the state value for updating the actor, and an evaluator to assess the evolved transition trajectories and select the leading action from the best trajectory.

\renewcommand{\algorithmicrequire}{\textbf{Input:}}
\renewcommand{\algorithmicensure}{\textbf{Output:}} 
\begin{algorithm}[!t]
\caption{Training EvoRL agent for puzzle reassembly}
\label{alg:ppo}
\begin{algorithmic}[1]
\REQUIRE Number of learning iterations $K$, maximum number of swaps $T$, number of evolution iterations $Z$, action population size $S$, memory buffer $\mathbb{B}$, training dataset $\mathbb{D}$\\
\ENSURE actor $\omega_\theta$, critic  $\omega_\varphi$, and evaluator $\omega_\mu$ 
\FOR{$k \gets 1$ \TO $K$}
    \STATE Randomly select a puzzle from $\mathbb{D}$
    \FOR{$t \gets 1$ \TO $T$}
        \STATE Sample a set of actions $\hat{\bm{a}}_{t}$ from the actor\\
        \STATE Combine the sampled actions with the historical actions as the initial parents\\
            \FOR{$z \gets 1$ \TO $Z$}
                \STATE Generate $S$ offspring by crossover and mutation\\
                \STATE Evaluate all offspring by using the evaluator $\omega_\mu$ and select new parents \\
            \ENDFOR
        \STATE Select the best-performing action ${a}_{t}$ from the evolved offspring by using the evaluator $\omega_\mu$\\
        \STATE Execute the action $a_t$ onto the puzzle and derive the reward $\mathcal{R}(s_t, {a}_t)$ as in Eq.~\eqref{eqn:reward}\\
        \STATE Store the trajectory $ \tau =  \{s_t, a_t, \mathcal{R}(s_t, a_t), s_{t+1} \}$ in memory buffer $\mathbb{B}$ 
    \ENDFOR
    \STATE Sample a batch of transitions $\tau$ from $\mathbb{B}$
    \STATE Calculate the state value $\mathcal{V}(s_t)$ as in Eq.~\eqref{eqn:vvalue} 
    \STATE Compute the advantage function $\mathcal{A}(a_t)$ as Eq.~\eqref{eqn:advantage}
    \STATE Update the actor $\omega_\theta$ using gradient ascent with the clipped objective function as in Eq.~\eqref{eqn:actor_gradient}
    \STATE Update the critic $\omega_\varphi$ by gradient descent as in Eq.~\eqref{eqn:critic_gradient} 
    \STATE Soft update the evaluator $\omega_\mu$ as in Eq.~\eqref{eqn:evaluator_update}
\ENDFOR
\end{algorithmic}
\end{algorithm}

The procedure for training the EvoRL agent is summarized in Algo.~\ref{alg:ppo}. Given a shuffled puzzle in state $s_t$, a set of swapping actions $\hat{\bm{a}}_t \in \mathbb{A}$ are sampled from the action space using the actor policy $\omega_\theta$ according to the probability $P(\hat{a}_t|s_t)$. The recommended actions $\hat{\bm{a}}_t$ and the historical actions from the replay buffer are concatenated as the initial parents. Then, the evolutionary operations such as crossover and mutation~\cite{mirjalili2019genetic} are applied iteratively to produce offsprings, and the evaluator $\omega_\mu$ assesses the offspring, and selects the action $a_t$ with the highest value as the optimal swapping action. The corresponding reward $\mathcal{R}(s_t, a_t)$ is calculated using Eq.~\eqref{eqn:reward}. After executing the action $a_t$ onto the current puzzle, the puzzle proceeds to the next state $s_{t+1}$. 
The trajectory $\tau =  \{s_t, a_t, \mathcal{R}(s_t, a_t), s_{t+1} \}$ is stored in the replay buffer $\mathbb{B}$, where each trajectory is a sequence of states, actions, rewards, and next states. These trajectories are later sampled from $\mathbb{B}$ as the past reassembly experience via the experience replay mechanism. 

To guide the training of the actor, the state value $\mathcal{V}(s_t)$ is estimated by the critic as in~\cite{sutton2018reinforcement},
\begin{align}
\mathcal{V}(s_t) = \mathbb{E}_{a_t,s_{t+1},\dots} \left[\sum^{\infty}_{l = 0} \gamma^l \mathcal{R}(s_{t+l}, a_{t+l})\right],
\label{eqn:vvalue}
\end{align} 
where $\mathbb{E}_{a_t,s_{t+1},\dots}$ denotes the expectation over all future feasible states, and $\gamma$ is the discount factor to avoid greedily maximizing the reward at the current state. For training stability, the advantage function is calculated,
\begin{align}
\mathcal{A}(s_t,a_t) = \mathcal{R}(s_t, a_t) + \gamma \mathcal{V}(s_{t+1}) - \mathcal{V}(s_t),
\label{eqn:advantage}
\end{align}
which evaluates whether the action $a_t$ is better than other possible actions in state $s_t$. A CLIP function $\mathcal{F}_{c}(\mathcal{A}(s_t,a_t), \epsilon)$ of the surrogate objective is introduced to constrain the update of $\mathcal{A}(s_t,a_t)$ by a factor of $[1-\epsilon, 1+\epsilon]$ as in~\cite{schulman2017proximal}. 
The actor is then updated by gradient ascent as, 
\begin{align}
\label{eqn:actor_gradient}
L_{p} = & \min[P(a_t|s_t)\cdot\mathcal{A}(s_t,a_t), \mathcal{F}_{c}(\mathcal{A}(s_t,a_t), \epsilon)]\\
\theta \leftarrow & \theta + \eta_{\theta} \cdot \nabla_\theta(L_{p}),
% \theta \leftarrow \theta+\eta_{\theta} \nabla_{\theta} \log \omega_\theta (a_t | s_t)\mathcal{A}(s_t, a_t),
\end{align}
where $\eta_{\theta}$ is the learning rate. 
Following \cite{schulman2017proximal}, the value loss $L_v$ is defined as the squared error of the estimated state value in the trajectory. The critic $\omega_\varphi$ is then updated by gradient descent as,
\begin{align}
\label{eqn:critic_gradient}
\varphi \leftarrow \varphi-\eta_{\varphi}\nabla_\varphi\left(\mathcal{V}\left(s_t\right)-\sum_{i=t}^T \gamma^{i-t} \mathcal{R}(s_i, a_i)\right)^2,    
\end{align}
where $\eta_{\varphi}$ is the learning rate. 
The evaluator assesses all the offspring and selects the optimal swapping action $a_t$ from the offspring. It shares the same architecture as the critic and it is soft-updated by the critic as, 
\begin{align}
\label{eqn:evaluator_update}
\mu \leftarrow \beta * \varphi + (1-\beta) * \mu,  
\end{align}
where $\beta$ is the soft update parameter. The soft-updating mechanism ensures the evaluator to approximate the state value function from the rewards to properly evaluate the offspring, and more importantly, it prevents the evaluator from overfitting caused by the hard update~\cite{wei2021deep}. 

During testing, the actor evaluates the visual perception from the MPPN and generates a sequence of swapping actions by using the evolutionary strategy with the assistance from the evaluator till it perfectly assembles the puzzle.

\section{Experimental Results}
\label{sec: experiments}

\subsection{Experimental Settings}
\label{sec: experimental settings}
The proposed ERL-MPP is compared with Deepzzle~\cite{paumard2020deepzzle}, Greedy Search~\cite{paikin2015solving}, Tabu Search~\cite{adamczewski2015discrete}, Genetic Algorithm~\cite{sholomon2013genetic}, Inpaint-GAN~\cite{bridger2020solving}, PDN-GA~\cite{song2023solving} 
and SD$^2$RL~\cite{song2023sd2rl} on the following two benchmark datasets. 

\noindent \textbf{JPLEG-5 Dataset}~\cite{song2023sd2rl} contains 12,000 puzzles, where each puzzle of $534\times534$ pixels is divided into $5\times5$ random shuffled pieces with 12-pixel gaps. Following the same experimental setup as in SD$^2$RL~\cite{song2023sd2rl}, 9,000 puzzles are selected as the training set, 1,000 as the evaluation set, and 2,000 as the test set. 

\noindent \textbf{MIT Dataset}~\cite{cho2010probabilistic} consists of 60 puzzles. Each puzzle is resized and cut into either $7 \times 10$ pieces or $10\times 15$ pieces, and each piece has $64 \times 64$ pixels. Two types of gaps are applied, 2-pixel gap and 4-pixel gap. Following the same settings as in~\cite{bridger2020solving}, the models are trained on the DIV2K dataset~\cite{agustsson2017ntire} and tested on this dataset. 

For a fair comparison, we adopt the same evaluation metrics as in~\cite{song2023sd2rl}, \emph{i.e.}, \textbf{Perfect}, \textbf{Absolute}, \textbf{Horizontal} and \textbf{Vertical}, showing the percentage of puzzles that are perfectly reassembled, in their correct absolute positions, in correct horizontal
and vertical pairwise relations, respectively. The last two are often combined as \textbf{Neighbor} metric. 

The hyper-parameters for the proposed ERL-MPP are set as follows: the discount rate $\gamma = 0.998$, the buffer size 100, the maximum number of iterations $K = 1000$, the maximum number of swaps $T=10,000$, the number of evolution iterations $Z=10$, the action population size $S=64$, $\alpha =  0.8$ for Absolute reward, $b = 1$, and $r^P_t = 1000$ for a perfect reassembly. The Adam optimizer is used. More details are provided in the supplementary material.

\subsection{Comparison Results on JPLEG-5 Dataset}
\label{JPLEG-5 Evaluation}
The comparison results on the JPLEG-5 dataset are summarized in Table~\ref{table:JPLEG-5}. It is indeed difficult to reassemble puzzles with large gaps. The two previous best performing methods, PDN-GA and SD$^2$RL, only achieve a \textbf{Perfect} rate of 6.1\% and 5.1\% respectively, and others almost fail to perfectly reassemble any puzzles. The proposed method obtains a \textbf{Perfect} rate of $18.6\%$, significantly better than all the compared methods. Compared to the previous best performing method, PDN-GA~\cite{song2023solving}, the performance gains are $12.5\%$, $8.4\%$, $25.7\%$, and $26.7\%$ in terms of \textbf{Perfect}, \textbf{Absolute}, \textbf{Horizontal}, and \textbf{Vertical} metrics, respectively.

\begin{table}[!t]
\begin{center}
\begin{tabular}{|l|c|c|c|c|}
    \hline
    Method & \textbf{Perf.} & \textbf{Abs.} & \textbf{Hori.} & \textbf{Vert.} \\
    \hline
    Deepzzle~\shortcite{paumard2020deepzzle} & 0.0 & 21.9 & 10.9 & 10.7 \\
    \hline
    Greedy~\shortcite{paikin2015solving} & 0.1 & 24.1 & 12.6 & 12.3 \\
    \hline
    Tabu~\shortcite{adamczewski2015discrete} & 0.0 & 24.6 & 12.8 & 12.8\\
    \hline
    GA~\shortcite{sholomon2013genetic} & 0.0 & 25.1 & 12.4 & 12.3\\
    \hline
    SD$^2$RL~\shortcite{song2023sd2rl} & 5.1 & 40.3 & 26.5 & 26.2 \\
    \hline
    PDN-GA~\shortcite{song2023solving} & \underline{6.1} & \underline{44.3} & \underline{30.8} & \underline{30.6} \\
    \hline
    Proposed ERL-MPP & \textbf{18.6} & \textbf{52.7} & \textbf{56.5} & \textbf{57.3}\\
    \hline
\end{tabular}
\end{center}
\caption{ERL-MPP significantly outperforms the compared methods in terms of four metrics on the JPLEG-5 dataset.}
\label{table:JPLEG-5}
\end{table}

The JPLEG-5 dataset consists of three main types of puzzles: paintings (Pnt.), engravings (Eng.), and artifacts (Art.). The evaluation results on these three types of puzzles are summarized in Table \ref{table:3}. The proposed ERL-MPP significantly outperforms all the compared methods on puzzles of different types using both metrics. Compared to the second best method PDN-GA~\cite{song2023solving}, the performance gains are 2.7\%, 15.8\%, and 19.0\%, respectively on painting, engraving, and artifact puzzles in terms of \textbf{Perfect} metric, and 4.0\%, 8.3\%, 12.8\% in terms of \textbf{Absolute} metric. The painting puzzles are relatively more challenging because of diverse image contents. Thanks to its superior ability to explore and exploit the large action space, the ERL-MPP correctly reassembles many more puzzles than others.

\begin{table}[!t]
\begin{center}
\begin{tabular}{|l|c|c|c|}
    \hline
    Method & Pnt. & Eng. & Art. \\
    \hline
    Deepzzle~\shortcite{paumard2020deepzzle} & 0.0/15.5 & 0.0/23.9 & 0.0/26.3 \\
    \hline
    Greedy~\shortcite{paikin2015solving} & 0.0/16.2 & 0.0/27.3 & 0.3/28.7 \\
    \hline
    Tabu~\shortcite{adamczewski2015discrete} & 0.0/16.4 & 0.0/28.4 & 0.0/29.0 \\
    \hline
    GA~\shortcite{sholomon2013genetic} & 0.0/17.0 & 0.0/28.9 & 0.0/29.4 \\
    \hline
    SD$^2$RL~\shortcite{song2023sd2rl} & 0.5/23.8 & 7.6/48.5 & 7.2/48.6\\
    \hline
    PDN-GA~\shortcite{song2023solving} & \underline{0.6}/\underline{24.5} & \underline{8.6}/\underline{54.5} & \underline{9.1}/\underline{54.0}\\ 
    \hline
    Proposed ERL-MPP & \textbf{3.3}/\textbf{28.5} & \textbf{24.4}/\textbf{62.8} & \textbf{28.1}/\textbf{66.8}\\ 
    \hline
\end{tabular}
%}
\end{center}
\caption{Comparison results on the JPLEG-5 dataset in terms of \textbf{Perfect} (left) and \textbf{Absolute} (right) metrics. The proposed ERL-MPP significantly outperforms all compared models.}
\label{table:3}
\end{table}

\begin{table*}[!hbtp]
    \begin{center}
    \begin{tabular}{|l|c|c|c|c|c|c|c|c|c|c|c|c|}
        \hline
        \multirow{3}*{Method} & \multicolumn{6}{c|}{$7\times10$ Pieces} & \multicolumn{6}{c|}{$10\times15$ Pieces} \\
        \cline{2-13}        
        & \multicolumn{3}{c|}{2-pixel gap } & \multicolumn{3}{c|}{4-pixel gap } & \multicolumn{3}{c|}{2-pixel gap } & \multicolumn{3}{c|}{4-pixel gap } \\
        \cline{2-13}
        & \textbf{Perf.} & \textbf{Abs.} & \textbf{Neig.} & \textbf{Perf.} & \textbf{Abs.} & \textbf{Neig.} & \textbf{Perf.} & \textbf{Abs.} & \textbf{Neig.} & \textbf{Perf.} & \textbf{Abs.} & \textbf{Neig.} \\
        \hline
        Greedy~\shortcite{paikin2015solving} & 1.7 & 42.9 & 68.4 & 0.0 & 12.1 & 41.4 & 0.0 & 42.2 & 65.9 & 0.0 & 11.1 & 39.7 \\
        \hline
        Tabu~\shortcite{adamczewski2015discrete} & 3.3 & 49.6 & 72.3 & 0.0 & 20.2 & 45.8 & 0.0 & 48.7 & 67.3 & 0.0 & 12.4 & 40.3 \\
        \hline
        GA~\shortcite{sholomon2013genetic} & 3.3 & 50.1 & 72.7 & 0.0 & 20.1 & 46.2 & 0.0 & 49.3 & 69.4 & 0.0 & 12.5 & 40.5 \\
        \hline
        Inpaint-GAN~\shortcite{bridger2020solving} & 6.7 & 86.0 & 84.6 & 1.7 & 50.5 & 57.1 & 1.7 & 66.7 & 76.3 & 0.0 & 35.4 & 51.3 \\
        \hline
        PDN-GA~\shortcite{song2023solving} & 8.3 & 86.5 & 85.3 & 3.3 & 52.6 & 58.8 & 1.7 & 67.8 & 77.8 & 0.0 & 37.6 & 54.8 \\
        \hline
        SD$^2$RL~\shortcite{song2023sd2rl} & \underline{11.7} & \underline{86.7} & \underline{85.4} & \underline{5.0} & \underline{53.4} & \underline{59.2} & \underline{3.3} & \underline{68.1} & \underline{78.0} & 0.0 & \underline{38.3} & \underline{55.3}\\ 
        \hline
        Proposed ERL-MPP & \textbf{50.0} & \textbf{90.8} & \textbf{89.2} & \textbf{13.3} & \textbf{59.4} & \textbf{67.3} & \textbf{8.3} & \textbf{70.9} & \textbf{80.4} & 0.0 & \textbf{40.9} & \textbf{59.7} \\
        \hline
    \end{tabular}
    \end{center}
    \caption{Comparison results on the MIT dataset with $7\times 10$ and $10\times 15$ pieces. The proposed method outperforms all the compared methods in terms of all three evaluation metrics for puzzles of different pieces with different gaps.}
    \label{table:MIT_70_150}
\end{table*}

\begin{figure}[!t]
    \centering
    \includegraphics[width=1.\linewidth]{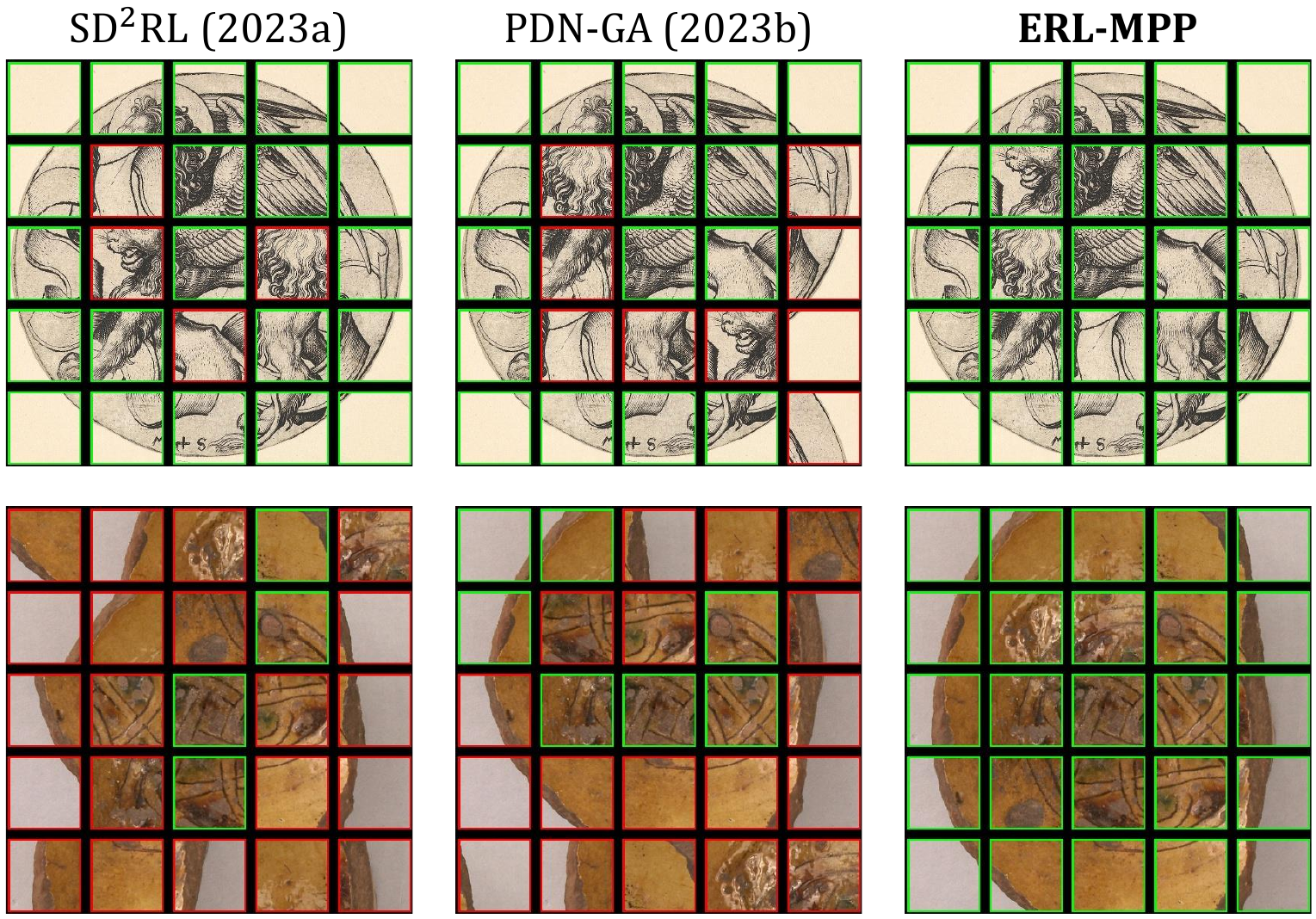}
    \caption{Visualization of the reassembling results on sample puzzles from the JPLEG-5 dataset~\cite{song2023sd2rl}.} %\rjf{Add references to compared methods.}}
    \label{fig:sample_results}
    % \vspace{-10pt}
\end{figure}

The reassembling results of top-3 performing models are shown in Fig. \ref{fig:sample_results}. The compared methods often make mistakes for pieces at core areas, while the proposed method could derive the correct reassembly order. The performance gain is mainly attributed to the excellent visual perception capabilities of the MPPN for understanding the image semantics and the capabilities of efficiently and effectively exploring the large action space by the EvoRL agent.

\subsection{Comparison Results on MIT Dataset}
\label{sec: MIT evaluation} 
To evaluate the models on large-scale puzzles,  we conduct comparison experiments on the MIT dataset. We adopt four experimental settings, puzzles of $7\times10$ and $10\times15$ pieces with 2-pixel and 4-pixel gaps, respectively. The results are summarized in Table \ref{table:MIT_70_150}. The proposed method achieves the best performance among all the compared methods in terms of three evaluation metrics under all four settings. Specifically, for $7\times10$ puzzles, compared to the second best method SD$^2$RL~\cite{song2023sd2rl}, ERL-MPP achieves a \textbf{Perfect} rate of 50.0\% and 13.3\% for 2-pixel gap and 4-pixel gap respectively, approximately as 4 and 3 times high as that of SD$^2$RL. With larger 4-pixel gaps, the performance of all the methods significantly decreases compared to that of 2-pixel gap, showing the difficulty of reassembling large-scale puzzles of large gaps. For $10\times15$ puzzles, ERL-MPP doubles the perfect reassembled puzzles with 2-pixel gap compared to SD$^2$RL, while none of the methods perfectly assemble any puzzles of 4-pixel gaps. Comparing the results for $7\times10$ puzzles and $10\times15$ puzzles, we can observe that the difficulty of puzzle reassembly increases dramatically with the number of puzzle pieces, \eg, for 2-pixel gaps, the \textbf{Perfect} metric for the proposed ERL-MPP decreases from 50.0\% to 8.3\% only when the number of puzzle pieces increases from 70 to 150.  Dividing a puzzle into too many small pieces with gaps will result in too little semantic information in each fragment and hence no method could work well in this case. Despite all the challenges, the proposed ERL-MPP significantly outperforms all the compared methods in terms of all the three metrics under all the four settings.

\begin{figure}[!t]
    \centering
    \includegraphics[width=1.\linewidth]{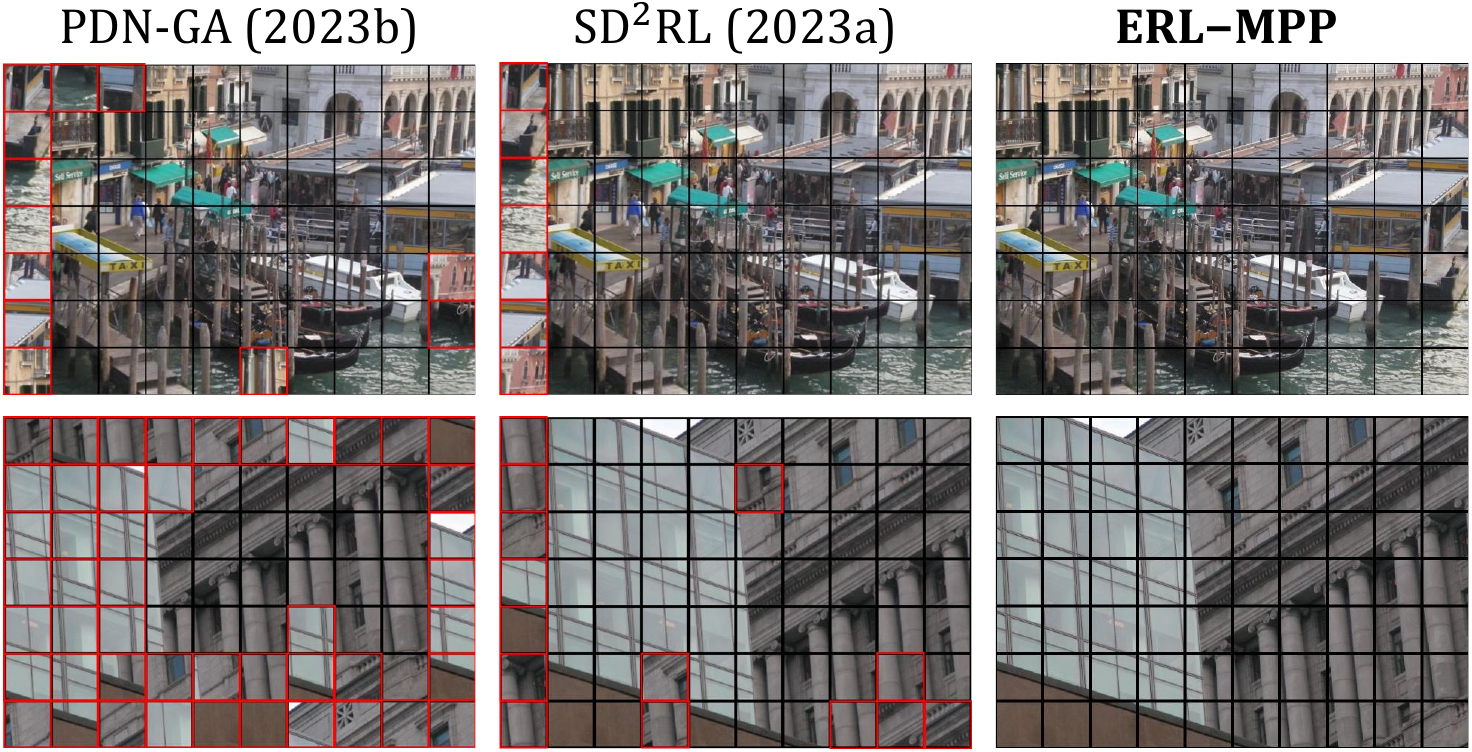}
    \caption{Visualization of the reassembled puzzles of $7\times 10$ pieces with 2-pixel gaps and 4-pixel gaps in the MIT dataset.}
    \label{fig:sample_results_70}
\end{figure}

Fig.~\ref{fig:sample_results_70} visualizes the final puzzles obtained by top-3 models. Compared to the perfect solutions by the ERL-MPP, most puzzle pieces are well reassembled by SD$^2$RL~\cite{song2023sd2rl} while some incorrect pieces ruin the solution. In contrast, PDN-GA~\cite{song2023solving} does not well reassemble the two puzzles with many pieces wrongly placed. The results demonstrate the power of ERL-MPP in puzzle reassembly, particularly the importance of evolutionary strategies that efficiently explore more swapping actions and enable the agent to escape from local optima. 

\subsection{Ablation Study}
\label{sec: ablation-study}
We ablate two key contributions: MPPN and EvoRL on the JPLEG-5 dataset.  SD$^2$RL~\cite{song2023sd2rl} is selected as the baseline, which utilizes the Siamese Discriminant Network (SDN) for puzzle perception and the Deep Q-Network (DQN) for puzzle reassembly. Table \ref{table:abs-JPLEG-5} presents the ablation results by gradually replacing SDN and DQN by MPPN and EvoRL. Both MPPN and EvoRL bring significant gains. The performance gain of the MPPN over the SDN is 5.1\%, 2.5\%, 12.7\%, and 13.5\% in terms of the \textbf{Perfect}, \textbf{Absolute}, \textbf{Horizontal}, and \textbf{Vertical} metrics, respectively when using the EvoRL agent, demonstrating the power of the shared-encoder and multi-head perception in extracting both local and global puzzle information. 
Compared to the DQN, the EvoRL achieves the performance gain of 8.8\%, 6.0\%, 18.7\%, and 19.6\% respectively when using the MPPN for visual perception, demonstrating the effectiveness of the EvoRL agent in exploring and exploiting the large action space. 
The ablation results confirm the contributions of both core components to the excellent performance. 

\begin{table}[!t]
\begin{center}
\resizebox{1.0\columnwidth}{!}{
\begin{tabular}{|c|c|c|c|c|c|}
    \hline
    \textbf{Perception} & \textbf{Agent} &\textbf{Perf.} & \textbf{Abs.} & \textbf{Hori.} & \textbf{Vert.} \\
    \hline
    SDN & DQN & 5.1 & 40.3 & 26.5 & 26.2\\
    \hline
    MPPN & DQN & 9.8 & 46.7 & 37.8 & 37.7 \\
    \hline
    SDN & EvoRL & 13.5 & 50.2 & 43.8 & 43.8 \\
    \hline
    MPPN & EvoRL & \textbf{18.6} & \textbf{52.7} & \textbf{56.5} & \textbf{57.3} \\
    \hline
\end{tabular} }
\end{center}
\caption{Ablation study of major components of the ERL-MPP on the JPLEG-5 dataset~\cite{song2023sd2rl}.}
\label{table:abs-JPLEG-5}
\end{table}

\section{Conclusion}
\label{sec: conclusion}
The proposed ERL-MPP tackles the challenges of solving L-JPEG problems. Specifically, 
the proposed MPPN visually perceives both the global semantics through the discriminator head and the local puzzle status through the three puzzlet perception heads, which provides informative visual clues for the EvoRL agent. To tackle the challenges of reassembling a large number of puzzle fragments with gaps, the proposed EvoRL agent selects a sequence of swapping actions based on both the current visual perception and past reassembly experience embedded in the agent. The novel actor-critic-evaluator architecture in conjunction with the evolutionary strategy demonstrates superior performance and efficiently explores and exploits the large action space featured with complex landscapes. Extensive experiments on puzzles with large gaps in the JPLEG-5 dataset and puzzles with large numbers of puzzle pieces with gaps in the MIT dataset show that the proposed ERL-MPP significantly outperforms all the compared methods in terms of all evaluation metrics under various scale and gap settings. 

% \clearpage
\newpage

\bibliography{main}

\end{document}